\def\year{2021}\relax
\documentclass[letterpaper]{article} 
\usepackage{aaai21}  
\usepackage{times}  
\usepackage{helvet} 
\usepackage{courier}  
\usepackage[hyphens]{url}  
\usepackage{amsmath}
\usepackage{booktabs}
\usepackage{amssymb}
\usepackage{verbatim}
\usepackage{graphicx} 
\usepackage{multirow}
\usepackage{natbib}  
\usepackage{caption} 
\usepackage[switch]{lineno}

\newcommand{\etal}{\textit{et al}.~}

\newcommand{\ieno}{\textit{i}.\textit{e}.}
\newcommand{\etcno}{\textit{etc}.}

\newcommand{\egno}{\textit{e}.\textit{g}.}
\usepackage{color}

\nocopyright
\title{Learning Omni-frequency Region-adaptive Representations for Real Image Super-Resolution}
\author{
    Xin Li\textsuperscript{\rm 1}\thanks{The first two authors contribute equally to this work.},
    Xin Jin\textsuperscript{\rm 1}\footnotemark[1], 
    Tao Yu\textsuperscript{\rm 1},
    Yingxue Pang\textsuperscript{\rm 1},
    Simeng Sun\textsuperscript{\rm 1},
    Zhizheng Zhang\textsuperscript{\rm 1},
    Zhibo Chen\textsuperscript{\rm 1}\thanks{Corresponding author.} \\
}
\affiliations{
    \textsuperscript{\rm 1}CAS Key Laboratory of Technology in Geo-spatial Information Processing and Application System,\\
University of Science and Technology of China \\
Hefei 230027, China\\

    \{lixin666, jinxustc, yutao666, pangyx, smsun20, zhizheng\}@mail.ustc.edu.cn, chenzhibo@ustc.edu.cn

}

\begin{document}

\maketitle

\begin{abstract}

Traditional single image super-resolution (SISR) methods that focus on solving \emph{single} and \emph{uniform} degradation (\ieno, bicubic down-sampling), typically suffer from poor performance when applied into real-world low-resolution (LR) images due to the complicated realistic degradations. The key to solving this more challenging real image super-resolution (RealSR) problem lies in learning feature representations that are both \emph{informative} and \emph{content-aware}. In this paper, we propose an \textbf{O}mni-frequency \textbf{R}egion-adaptive \textbf{Net}work (OR-Net) to address both challenges, here we call features of all low, middle and high frequencies omni-frequency features. Specifically, we start from the frequency perspective and design a Frequency Decomposition (FD) module to separate different frequency components to comprehensively compensate the information lost for real LR image. Then, considering the different regions of real LR image have different frequency information lost, we further design a Region-adaptive Frequency Aggregation (RFA) module by leveraging dynamic convolution and spatial attention to adaptively restore frequency components for different regions. The extensive experiments endorse the  effective, and scenario-agnostic nature of our OR-Net for RealSR.

\end{abstract}

\section{Introduction}

With the development of deep learning, single image super-resolution (SISR) has achieved great success either on PSNR values ~\cite{dong2015image, haris2018deep,kim2016accurate,lim2017enhanced,zhang2018image,dai2019second,Mei_2020_CVPR, pan2020image} or on visual quality~\cite{ledig2017photo,sajjadi2017enhancenet}. In general, these traditional SISR methods typically focus on restoring the low-resolution (LR) image with \emph{single} and \emph{uniform} synthetic degradation, such as bicubic down-sampling and Gaussian down-sampling. However, the degradations in real-world LR images are usually far more complicated, which makes most SISR models become less effective when directly applied to practical scenarios.

In recent years, some studies that focus on solving the real image super-resolution (RealSR) problem have attracted more and more attention~\cite{cai2019toward, wei2020component}. Unlike the \emph{single} and \emph{uniform} synthetic degradation in SISR, the LR and HR images of RealSR are captured with digital single lens reflex (DSLR) cameras, which typically contains various/complex non-uniform real-world degradations, including blur, noise, and down-sampling. That is why those classic conventional SISR methods (RCAN \cite{zhang2018image}, EDSR \cite{lim2017enhanced}, SAN \cite{dai2019second} \etcno) cannot handle the RealSR problem well.  
 
\begin{figure}
	\centering
	\includegraphics[width=\linewidth]{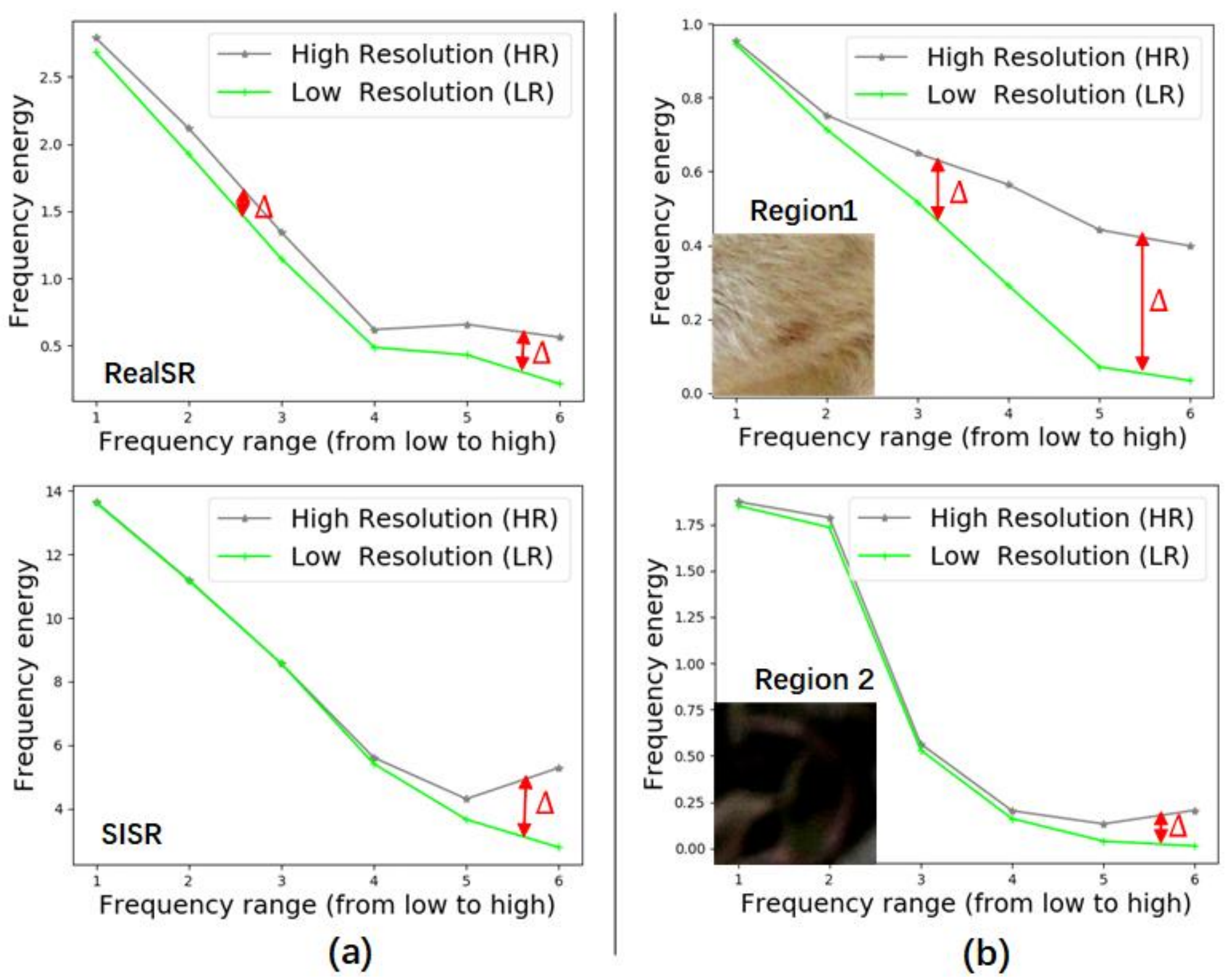}
	\vspace{-6mm}
	\caption{Difference analysis in frequency domain between RealSR and the conventional SISR: (a) the overall frequency distribution comparison between RealSR and SISR. (b) the frequency components distribution comparison between different regions of the same real LR image. $\Delta$ means the absolute difference of frequency degradation.}
	\label{fig:analysis_motivation}
	\vspace{-4mm}
\end{figure}

Among recent representative RealSR methods, Cai \etal~\cite{cai2019toward} first construct a RealSR dataset by capturing the LR-HR image pairs on the same scene with different focal lengths of a digital camera. And then, they propose a Laplacian pyramid based kernel prediction network (LP-KPN) to handle such non-uniform degradation. Concurrently, Wei \etal~\cite{wei2020component} also propose a large-scale diverse real-world image super-resolution dataset named DRealSR. Considering the targets of RealSR vary with image regions, they further design a component divide-and-conquer (CDC) model to adaptively restore the LR image. But, LP-KPN and CDC both only focus on super-resolving LR images by collaborating different pixel-wise local restorations (\egno, flat regions, edges and corners), they ignore to make full use of hierarchical features across different frequency domains to comprehensively enhance texture details for real LR images. 




In this paper, we first analyse the latent and essential challenges of RealSR from a completely different perspective of frequency distributions. In particular, we perform an experimental comparison between RealSR and conventional SISR in the frequency domain to explore their difference. Specifically, we visualize two LR-HR pairs in frequency domain through the wavelet transform tool \cite{rao2002wavelet}, where one pair belongs to SISR and the other belongs to RealSR. As shown in Figure \ref{fig:analysis_motivation}(a), we observe that the degradation (from HR image to LR image) of the general SISR mainly exists in the high-frequency component. In contrast, the degradation of RealSR exists in all frequency components. Besides, in Figure~\ref{fig:analysis_motivation}(b), we also see that the degradation of different regions of an image are usually distributed in the different frequency components.

Based on the above analysis, we argue that an effective RealSR model should learn feature representations that are both \emph{informative} and \emph{content-aware}. \emph{Informative} property promises to restore sufficient realistic texture details across multiple frequency domains for the degraded LR, and \emph{content-aware} property satisfies the varied targets of RealSR for different image regions (\ieno, smoothing for flat regions and sharpening for edges). In this paper, we propose a \textbf{O}mni-frequency \textbf{R}egion-adaptive \textbf{Net}work (OR-Net) to efficiently solve the RealSR problem. Specifically, we first design a Frequency Decomposition (FD) module to decompose LR image into low-frequency, middle frequency , and high frequency components. Then, we employ multiple interactive branches to enhance the corresponding frequency factors. Second, to achieve the content-aware super-resolution for real LR images, we further design a Region-adaptive Frequency Aggregation (RFA) module by combining the dynamic convolution and spatial attention to selectively restore different frequency components for the different positions of HR images. The contributions of this paper can be summarized as follows:
\begin{itemize}
	\item We analyse the essential difference between generalSR and RealSR from the frequency perspective, to answer the question that why the classic SISR methods cannot handle RealSR problem well.
	
	\item Based on our analysis, we propose an \textbf{O}mni-frequency \textbf{R}egion-adaptive \textbf{Net}work (OR-Net) for RealSR, which contains two technical novelties--1) Frequency Decomposition (FD) module that aims to achieve the LR image content separation in frequency domain and enhance texture details across all frequency components, 2) Region-adaptive Frequency Aggregation (RFA) module that aims to appropriately restore different frequency components for real HR images in different positions.

	
	\item Extensive experiments on multiple RealSR benchmarks have validated the effectiveness and superiority of our OR-Net. Sufficient intuitive visualization results/analysis are also provided to support/verify the expected functions of the proposed FD and RFA modules.

\end{itemize}

\section{Related Work}
\subsection{Conventional Single Image Super-Resolution}

In the last decade, the traditional SISR has achieved great progress, especially for deep learning based approaches \cite{dong2015image,dong2016accelerating,lim2017enhanced,haris2018deep,zhang2018image,dai2019second}. These methods usually perform well on the the synthetic degradation (\egno, bicubic down-sampling) but generalize poorly to realistic complicated distortions in real-world scenarios. This is problematic especially in practical applications, where the target scenes typically have hybrid/complex non-uniform degradations (\egno, blur, noise, and down-sampling), and also, there is always no readily available paired data for training.


\subsection{Real Image Super-Resolution}
RealSR has drawn more and more attention in recent years. Different from the general SISR that typically focuses on the simple and uniform synthetic degradation, RealSR mainly aims to solve these realistic complicated degradations in real-world scenarios. To capture the distortion of real scene, Chen \etal~\cite{chen2019camera} design two novel data acquisition strategies. Cai \etal~\cite{cai2019toward} build a real-world super-resolution (RealSR) dataset by adjusting the focal length of a digital camera, and introduce the Laplacian pyramid based kernel prediction network (LP-KPN) to solve such non-uniform distortions. Recently, Wei \etal~\cite{wei2020component} present a large-scale diverse real-world image super-resolution dataset (DRealSR) and a component divide-and-conquer (CDC) model with gradient weighted (GW) loss, achieving great performance in RealSR. 

However, above methods ignore to consider and study the difference between general SISR and RealSR. They didn't design solutions based on the essential difference analysis, which is sub-optimal and un-targeted. In this paper, we start from analysing the difference between two kinds of super-resolution tasks in frequency domain in detail. Besides, we also analyse the degradation of different regions in the same LR image. Based on these analysis, we study how to design a generalizable and efficient RealSR framework that can exploit the merits of previous works while more targeting on the specific characteristics of RealSR itself.



\begin{figure*}[htp]
	\centering
	\includegraphics[width=\textwidth]{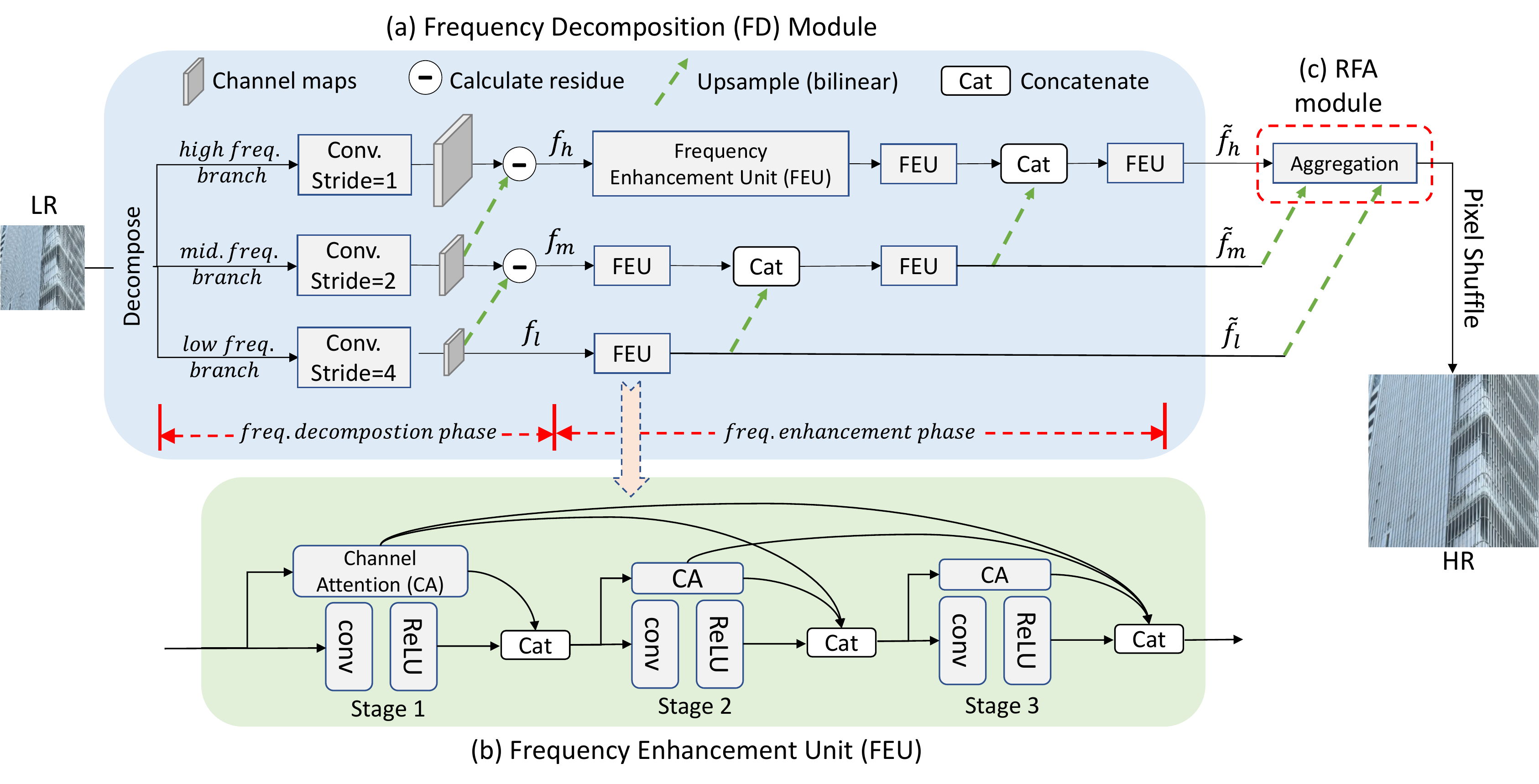}
	\vspace{-6mm}
	\caption{Architecture of our proposed OR-Net, which consists of two critical modules: (a) Frequency Decomposition (FD) module, which employs an multiple-branch architecture to decompose the input real LR image content in frequency domain and enhance texture details across all frequency components, (c) Region-adaptive Frequency Aggregation (RFA) module, which aggregates the enhanced omni-frequency components for different regions of a LR image. Moreover, in FD module, we additionally design a (b) Frequency Enhancement Unit (FEU) to strengthen the feature representation capability.}
	\label{fig:framework}
	\vspace{-4mm}
\end{figure*}

\subsection{Frequency Decomposition and Content Adaptation}

Lots of excellent low-level restoration studies explore to enhance/reconstruct content details from the frequency decomposition perspective, including image denoising/deraining~\cite{fu2017clearing}, rescaling~\cite{xiao2020invertible} and super-resolution \cite{fritsche2019frequency, pang2020fan}. Particularly, Chen \etal introduce an octave convolution~\cite{chen2019drop} to decompose features in the frequency domain to reduce spatial redundancy in CNNs. Akbari \etal~\cite{akbari2020generalized} further extend the octave convolution and frequency decomposition idea to the image compression field. These methods typically decompose the degraded image into low-/high-frequency factors, and individually deal with different components to achieve divide-and-conquer. However, they usually ignore the interaction and aggregation between multiple frequency factors. 
In addition, with the development of some content-adaptation techniques (\egno, attention mechanism~\cite{wang2018non,woo2018cbam,li2019selective,hou2020strip, Mei_2020_CVPR,zhang2020relation}, region normalization~\cite{yu2020region} and dynamic convolution~\cite{jia2016dynamic, niklaus2017video, mildenhall2018burst, lin2019context, chen2020dynamic}), some low-level restoration algorithms (\egno, image interpolation \cite{niklaus2017video}, hybrid-distorted image restoration \cite{li2020learning} and denoising \cite{mildenhall2018burst}) also achieve the purpose of `divide-and-conquer', where the processing operations can be adjusted adaptively according to the changing image content in the reference. In this paper, considering that the different regions of real LR image inevitably have different frequency information lost, we explore to simultaneously utilize dynamic convolution and spatial attention to achieve adaptive frequency information compensation.

\section{Omni-frequency Region-adaptive Network}

\begin{figure}[htp]
	\centering
	\includegraphics[width=0.95\linewidth]{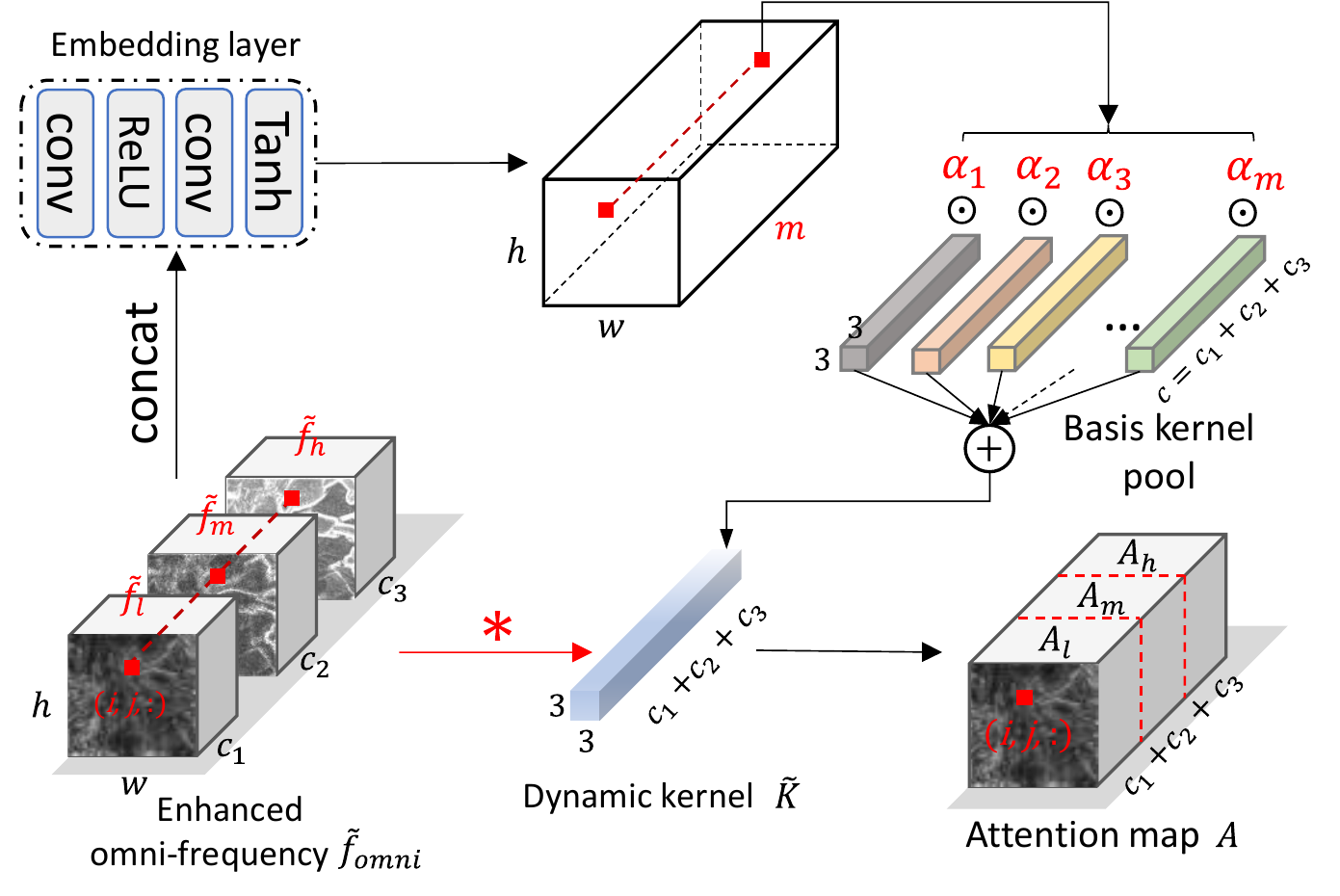}
	\caption{Region-adaptive attention map systhesis in RFA module. $*$ is matrix multiplication and $\odot$ is element multiplication.}
	\vspace{-3mm}
	\label{fig:RFAmodule}
\end{figure}

We aim at designing a generalized and efficient framework for RealSR. During the training, we have access to the annotated real-world SR datasets with LR-HR pairs. The trained model is expected to work well with high generalization capability for the real LR images. Figure \ref{fig:framework} shows the overall flowchart of our framework. Particularly, we propose an \textbf{O}mni-frequency \textbf{R}egion-adaptive \textbf{Net}work (OR-Net) to boost the SR performance on the real low-resolution images. OR-Net is designed from a frequency perspective, and contains two novel technical designs: Frequency Decomposition (FD) module and Region-adaptive Frequency Aggregation (RFA) module. As shown in Figure \ref{fig:framework}(a), FD module first employs an architecture with three branches, which decomposes the input real LR image into low-/middle-/high-frequency components (\ieno, omni-frequency) in frequency domain and enhance them in an interactive manner to achieve the comprehensive information compensation. Moreover, the RFA module (see Figure \ref{fig:framework}(c), \ref{fig:RFAmodule}) adaptively fuses the enhanced omni-frequency for different regions to achieve the content-aware super-resolution.

\subsection{Frequency Decomposition (FD) Module}
To obtain the informative omni-frequency representation for RealSR, we propose the \textbf{F}requency \textbf{D}ecomposition (FD) module. As shown in Figure \ref{fig:framework}(a), FD module is consists of two phases: frequency decomposition phase and frequency enhancement phase. The first frequency decomposition phase aims to separate the low-/middle-/high- frequency components from the LR input, and the frequency enhancement phase is dedicated to enhance the different frequency representations. To encourage the interaction between different frequency components and enhance frequency components in a coarse-to-fine (easy-to-hard) manner, we also progressively utilize the enhanced lower frequency representation to help the enhancement of higher frequency components by concatenating two frequency representations.

\begin{figure}[htp]
	\centering
	\includegraphics[width=\linewidth]{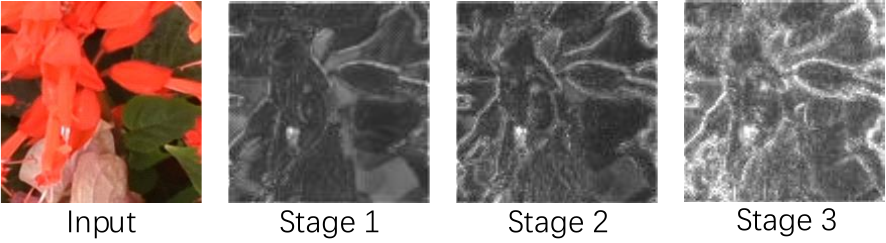}
	\caption{Feature visualization on the different stages of frequency enhancement unit (FEU).}
	\label{fig:dense}
\end{figure}
For the FD module, we denote the input (which is an RGB LR image) by $I \in \mathbb{R}^{h\times w \times 3}$ and the output includes three kinds of enhanced frequency features: low-frequency enhanced features $\widetilde{f}_l$, mid-frequency enhanced features $\widetilde{f}_m$, and high-frequency enhanced features $\widetilde{f}_h$.


\noindent\textbf{Frequency Decomposition Phase.} In FD module, we first decompose the input LR image $I$ into different frequency components. Such frequency components separation could be achieved in wavelet transform \cite{rao2002wavelet} or discrete consine transform \cite{ahmed1974discrete} in the traditional signal-processing methods. However,  with the mathematical operations being deterministic and task-irrelevant, such transforms inevitably discard some critical/detailed information for low-level restoration tasks. To imitate the wavelet transform while avoiding key information lost, we propose to factorizes the mixed feature representations through the \emph{learnable} latent-wise spatial down-sampling, the similar operation can be found in the recently-proposed octave convolution (OctConv)~\cite{akbari2020generalized} but OctConv aims to reduce channel-wise redundancy like group or depth-wise convolutions. Specifically, we first utilize the convolution layer with the larger stride (\egno, stride$=2$) to downsample the feature represent to extract the coarse features, \ieno, the low-frequency components. Then, we remove such relatively low frequency components from the original feature (before downsample) to obtain the rest relatively high frequency feature representations.

In formula, and as shown in Figure \ref{fig:framework}(a), we downsample the feature space by using convolution layer with stride$=4$ to get the corresponding low-frequency components $f_l$. Then we get the middle frequency components $f_m$ by removing the $f_l$ from the corresponding original features, which is also downsampled with stride$=2$. Similarly, to get the high frequency components $f_h$, we remove the downsampled features with stride$=2$ from the features without down-sampling, which has same spatial size with original LR image. The whole process can be denoted as follows:
\begin{equation}
\centering
\begin{split}
    & f_l=Conv{{\downarrow }_{2}}(Conv{{\downarrow }_{2}}(I)), \\ 
 & f_m=Conv{{\downarrow }_{2}}(I)-Conv{{\downarrow }_{2}}(Conv{{\downarrow }_{2}}(I)){{\uparrow }_{2}}, \\ 
 & f_h=Conv(I)-Conv{{\downarrow }_{2}}(I){{\uparrow }_{2}}, \\ 
\end{split}
\label{eq_fd}
\end{equation}
where $Conv{{\downarrow }_{2}}$ denotes the convolution layer with stride$=2$ and $Conv$ denotes the convolution layer without downsampling. $\uparrow$ means the bilinear upsampling operation. The corresponding interpretive/analysis experimental results that support the reasonableness of such frequency decomposition design can be found in Figure \ref{fig:fd_vis}.
\begin{table*}[htp]
\caption{Quantitative results on the DRealSR dataset. We compare our OR-Net to the general SISR methods, including \emph{Bicubic}, \emph{VDSR}, \emph{EDSR}, \emph{RDN}, \emph{DDBPN}, \emph{RCAN}, and RealSR methods, including \emph{LP-KPN} and \emph{CDC}. We use PSNR, SSIM and LPIPS as evaluation metrics.}
\label{tab:DRealSR}
\setlength{\tabcolsep}{1.6mm}{
\begin{tabular}{c|c|c|lll|c|lll|c|lll}
\toprule
\multirow{2}{*}{Method} & \multicolumn{1}{c|}{\multirow{2}{*}{Category}}&\multicolumn{1}{c|}{\multirow{2}{*}{Scale}} & \multicolumn{3}{c|}{DRealSR} & \multicolumn{1}{c|}{\multirow{2}{*}{Scale}} & \multicolumn{3}{c|}{DRealSR} & \multicolumn{1}{c|}{\multirow{2}{*}{Scale}} & \multicolumn{3}{c}{DRealSR} \\ \cline{4-6} \cline{8-10} \cline{12-14}
                     &   & \multicolumn{1}{l|}{}                       & PSNR    & SSIM    & \multicolumn{1}{l|}{LPIPS}  & \multicolumn{1}{l|}{}                       & PSNR    & SSIM    & \multicolumn{1}{l|}{LPIPS} & \multicolumn{1}{l|}{}                       & PSNR    & SSIM    & \multicolumn{1}{l}{LPIPS}   \\ \hline
Bicubic             &  \multirow{7}{*}{SISR}    & \multirow{7}{*}{$\times$2}                        & 32.67        & 0.877        &  0.201    & \multirow{7}{*}{$\times$3}                        & 31.50        & 0.835        & 0.362 & \multirow{7}{*}{$\times$4}                        & 30.56        & 0.820        & 0.438         \\   
VDSR       &            &                                             & 34.02        & 0.901        & 0.154    &        & 32.60        & 0.859        & 0.263  &    & 31.43        & 0.839        & 0.337      \\
EDSR       &           &                                             & 34.24        & 0.908        & 0.155   &                                             & 32.93        & 0.876        & 0.241    &                                             & 32.03        & 0.855        & 0.307 \\     
RDN       &            &                                             & 34.46        & 0.910        & 0.151  &                                             & 33.08        & 0.875        & 0.245     &                                             & 32.08        & 0.857        & 0.308  \\     
DDBPN    &              &                                             & 34.26        & 0.906        & 0.157   &    & - & - & -     &                                             & 31.80        & 0.849        & 0.321         \\    
RCAN    &                &                                             & 34.34        & 0.908        & 0.158   &                                             & 33.03        & 0.876        & 0.241     &                                             & 32.41        & 0.861        & 0.303         \\  \hline  
LP-KPN    &     \multirow{3}{*}{RealSR}          &    \multirow{3}{*}{$\times$2}                                           & 33.88        & -        &   -      &     \multirow{3}{*}{$\times$3}                                          & 32.64        & -        &  -    &        \multirow{3}{*}{$\times$4}                                       & 31.58        & -        &   -  \\  
CDC     &               &                                             & 34.45        & \textbf{0.910}        & 0.146    &                                             & 33.06        & 0.876        & 0.244           &                                             & 32.42        & 0.861        & 0.300 \\ 
OR-Net(Ours)    &       &                                             & \textbf{34.56}        & \textbf{0.910}        & \textbf{0.145}        &                                             & \textbf{33.28}        & \textbf{0.877}        & \textbf{0.230}   &                                             & \textbf{32.59}        & \textbf{0.863}        & \textbf{0.292}         \\ \bottomrule    

\end{tabular}}
\end{table*}

\noindent\textbf{Frequency Enhancement Phase.} 
After extracting the low-/middle-/high-frequency components from LR image, we enhance these representations by a well-designed Frequency Enhancement Unit (FEU), to make up for low-/middle-/high-frequency information lost. Specifically, the FEU is designed based on the popular GRDB module \cite{kim2019grdn}, which can be regarded as an dense connection block. But, as shown in the Figure \ref{fig:framework}(b) and Figure \ref{fig:dense}, the features on the different stages of FEU are usually different and contain different information (some features focus on object structure but others focus on texture details). Hence, the common fusion/concatenate (directly sum up all features) that used in traditional dense connection is not consistent with our purpose (\ieno, divide-and-conquer). Intuitively, the simple sum-up of all features cannot promise that the low-frequency branch only focuses on the low-frequency features, same issues also exist in other two branches (middle- and high-freq. branch). To address this problem, except of the regular operations (\egno, non-linear transformation achieved by Conv+ReLU, dense connection), we further integrate the channel-wise attention to adaptively adjust the residual information aggregation in FEU as shown in Figure \ref{fig:framework}(b), which helps each branch selectively fuse the corresponding frequency components at different stages, and thus the representation capability in frequency domain of each branch is significantly improved. 

Moreover, considering that the high-frequency feature components are relatively difficult to restore/enhance~\cite{fritsche2019frequency,wei2020component}, we propose to enhance such challenging frequency component in a coarse-to-fine/easy-to-hard manner. In detail, we encourage the interaction between different frequency components, and progressively utilize the enhanced lower frequency  representations to help the enhancement of higher frequency components by concatenating them together. We define this process as:
\begin{equation}
\centering
\begin{split}
    & \widetilde{f}_l=Enhance(f_l), \\
    & \widetilde{f}_m=Enhance(f_m, \widetilde{f}_l), \\ 
    & \widetilde{f}_h=Enhance(f_h, \widetilde{f}_m, \widetilde{f}_l), \\ 
\end{split}
\label{enhance}
\end{equation}
where $Enhance(\cdot)$ denotes a set of several frequency enhancement unit (FEU), the specific number of FEU of each branch can be found in Figure \ref{fig:framework}(a).

\vspace{-1mm}

\subsection{Region-adaptive Frequency Aggregation (RFA) Module}
Different from the conventional SISR, the degradation (from HR$\rightarrow$LR image) of RealSR generally exists in all frequency components. Besides, the frequency information loss of different regions in a real LR image is different. Therefore, it is necessary to adaptively aggregate omni-frequency components for different regions to restore a more realistic HR image with the rich texture details. In this section, we introduce the Region-adaptive Frequency Aggregation (RFA) module in detail (shown in Figure \ref{fig:RFAmodule})
, which achieves the content-aware super-resolution by combining the dynamic convolution and spatial attention mechanism. 


\begin{figure*}[h]
	\centering
	\includegraphics[width=0.85\textwidth]{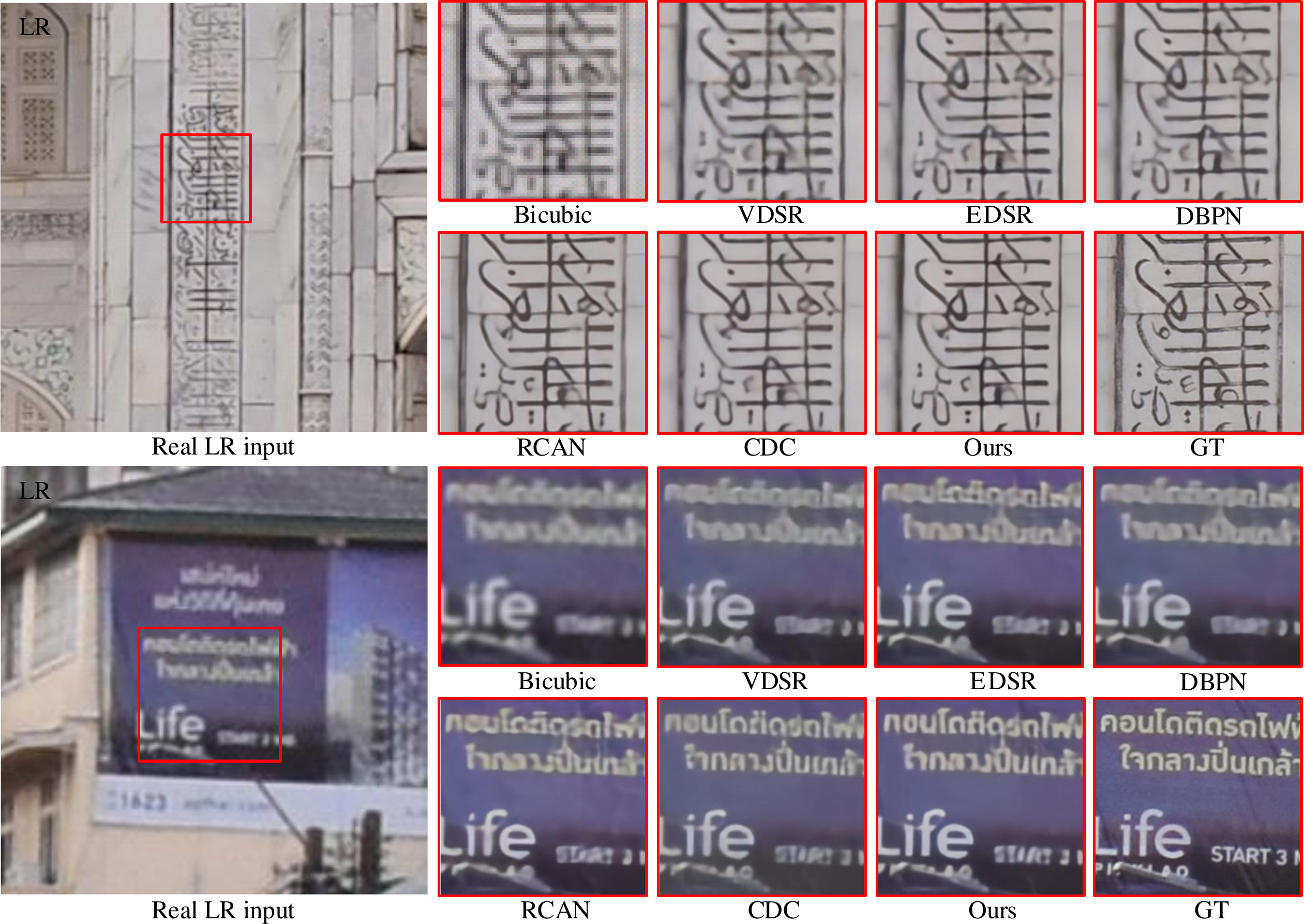}
	\caption{The qualitative comparison of our OR-Net with the state-of-the-art methods performed on DRealSR dataset ($\times$4 scale). The \emph{VDSR, EDSR, DBPN, RCAN} are designed for the general SISR and \emph{CDC} is specifically designed for RealSR. Note that all methods are re-trained on the DRealSR dataset for fairness.}
	\label{fig:vis_sub}
	\vspace{-2mm}
\end{figure*}

To achieve the region-adaptive aggregation, a straight-forward solution is to utilize spatial attention in \cite{woo2018cbam} to fuse the low-/mid-/high-frequency components. However, the general spatial attention only leverages the spatial contextual information but lacks of considering the relationship among low-/mid-/high frequency features, since the attention map for each region is generated through the same convolution filter. 
Therefore, in our RFA module, to achieve better omni-frequency feature adaptive aggregation, we combine the dynamic convolution \cite{zhao2018dynamic} and spatial attention \cite{woo2018cbam} to achieve adaptive frequency aggregation. Specifically, we utilize dynamic convolution kernels to synthesize the region-adaptive attention map from omni-frquency features. With such attention map, we achieve flexible and accurate frequency component fusion for different image regions.

As shown in Figure \ref{fig:RFAmodule}, we first concatenate the enhanced low-/mid-/high-frequency $\widetilde{f}_{l}$, $\widetilde{f}_{m}$ and $\widetilde{f}_{h}$ to get $\widetilde{f}_{omni}$:
\begin{equation}
\centering
    \widetilde{f}_{omni}=[({\widetilde{f}_{l}}{{\uparrow }_{2}}){{\uparrow }_{2}},({\widetilde{f}_{m}}){{\uparrow }_{2}},{\widetilde{f}_{h}}],
    \label{omni_f}
\end{equation}
where $[\cdot]$ represents the `concat' operation. Then we set up a learnable basis kernel pool $\mathcal{K} \in {{\mathbb{R}}^{m\times in\times c\times {{k}\times {{k}}}}}$, where  $m$, $in$, $c$, and ${k}$ represent the number of filters, input channel, output channel and kernel size (see Figure \ref{fig:RFAmodule}). After that, we pass $\widetilde{f}_{omni}$ through an embedding layer to obtain a coefficient tensor $\alpha \in {{\mathbb{R}}^{h\times w \times m}}$, where $h$ and $w$ represent the same height and width of omni-frequency  $\widetilde{f}_{omni}$. Finally, we re-weight $m$ filters in $\mathcal{K}$ with $\alpha (i,j,:)$ for region $(i, j, :)$ to get region-adaptive dynamic convolution filter $\widetilde{K}$:  
\begin{equation}
\centering
     \widetilde{K}=\sum\limits_{n=1}^{m}{\alpha_{n} (i,j,:){{K}_{n}}}.
    \label{weighted_sum}
\end{equation}


With dynamic convolution kernel $\widetilde{K}$, we can get the region-adaptive attention map $A(i, j, :)$ for region $(i,j,:)$ as:
\begin{equation}
\centering
\begin{split}
    & A(i, j, :)= \widetilde{f}_{omni}(i,j, :)*\widetilde{K}, \\
\end{split}
 \label{adjust}
\end{equation}
where $*$ represent the convolution operation. Finally, the aggregated omni-frequency feature $f$ can be obtained by:
\begin{equation}
\centering
\begin{split}
    & f(i, j, :) = A(i, j, :) \bullet  {\widetilde{f}_{omni}}, \\
\end{split}
 \label{fusion}
\end{equation}
where $\bullet$ represent the dot multiplication.

\vspace{-1mm}
\section{Experiments}
In this section, we first describe the datasets of RealSR and our implementation details in Section~\ref{sec:data_implementation}. And then, to verify the superiority of our method, we compare the proposed OR-Net with the current state-of-the-art RealSR methods and conventional SISR methods in Section~\ref{sec:SOTA}. To validate the effectiveness of the proposed FD and RFA modules, we show the visualization analysis in Section~\ref{sec:vis} and present a series of ablation studies for OR-Net in Section~\ref{sec:ab}.

\subsection{Dataset and Implementation Details}\label{sec:data_implementation}
RealSR is now under-explored and few works focus on such new challenge. Hence, these are few datasets can be used for evaluation. Here we evaluate our OR-Net on  DRealSR~\cite{wei2020component}.  DRealSR dataset is collected by \cite{wei2020component}. The training dataset contains 35,065, 26,118, and 30,502 image patches for scales of $\times$2, $\times$3 and $\times$4, respectively. The size for patches of scale $\times$2, $\times$3 and $\times$4 are 380$\times$380, 272$\times$272 and 192$\times$192. The testing dataset contains 83, 84, and 93 images for $\times$2 $\sim$ $\times$4, respectively.



The implementation of OR-Net is based on PyTorch framework. In the training process, we utilize Adam optimizer with an initial learning rate of 0.0001 and the learning rate decay by a factor of 0.5 each epoch. Batch size is 8 and we leverage random flip, random rotation and random cropping to achieve data augmentation. We randomly crop the training image as 192$\times$192. For FD module, we set the channels of three frequency branches as 128, 128 and 64 from low-frequency to high-frequency components. For RFA module, we set the number of basis kernels $K$ as 5.

$L_1$ loss has been verified effective and been widely used in many super-resolution works~\cite{lim2017enhanced, zhang2018image}. In this paper, we also utilize the $L_1$ loss to optimize our OR-Net.
\vspace{-2mm}
\subsection{Comparison with State-of-the-Arts}\label{sec:SOTA}


We compare our OR-Net with the state-of-the-art traditional SISR models (including \emph{Bicubic}, \emph{VDSR} \cite{kim2016accurate}, \emph{EDSR} \cite{lim2017enhanced}, \emph{RDN} \cite{zhang2018residual}, \emph{DDBPN} \cite{haris2018deep}, \emph{RCAN} \cite{zhang2018image}), and the RealSR models (including \emph{LP-KPN} \cite{cai2019toward} and \emph{CDC} \cite{wei2020component}). As shown in Table. \ref{tab:DRealSR}, our OR-Net achieves the best performance in terms of PSNR, SSIM and LPIPS compared to other general SISR and RealSR methods. We analyse that the general learning-based SISR methods that focus on solving synthetic degradation usually ignore the restoration of the full-frequency components of real LR, and cannot handle non-uniform distortions well. Besides, CDC and LP-KPN both ignore to leverage rich hierarchical information/features in frequency domains to comprehensively enhance texture details for real LR, which limits their practicality and scalability. In contrast, with the proposed frequency decomposition (FD) module and region-adaptive frequency aggregation (RFA) module, our OR-Net could enhance texture details for real LR images across all frequency scope while achieving content-aware super-resolution.

We provide the qualitative comparison of our OR-Net with state-of-the-art conventional SISR methods (including Bicubic, VDSR, EDSR, DBPN, and RCAN) and RealSR method (including CDC). We can see that those traditional SISR methods (including VDSR, EDSR, DBPN) fail to restore well some corrupted details of real LR, \egno, the letters in the second rows. RealSR method CDC cannot effectively remove blur artifacts. However, our OR-Net can achieve better restoration of textures and details for all regions.

We also evaluate our OR-Net on RealSR dataset~\cite{cai2019toward} and some traditional SISR datasets. More results can be found in \textbf{Supplementary}.

\vspace{-2mm}
\subsection{Visualization}\label{sec:vis}
\begin{figure}[htp]
	\centering
	\includegraphics[width=00.95\linewidth]{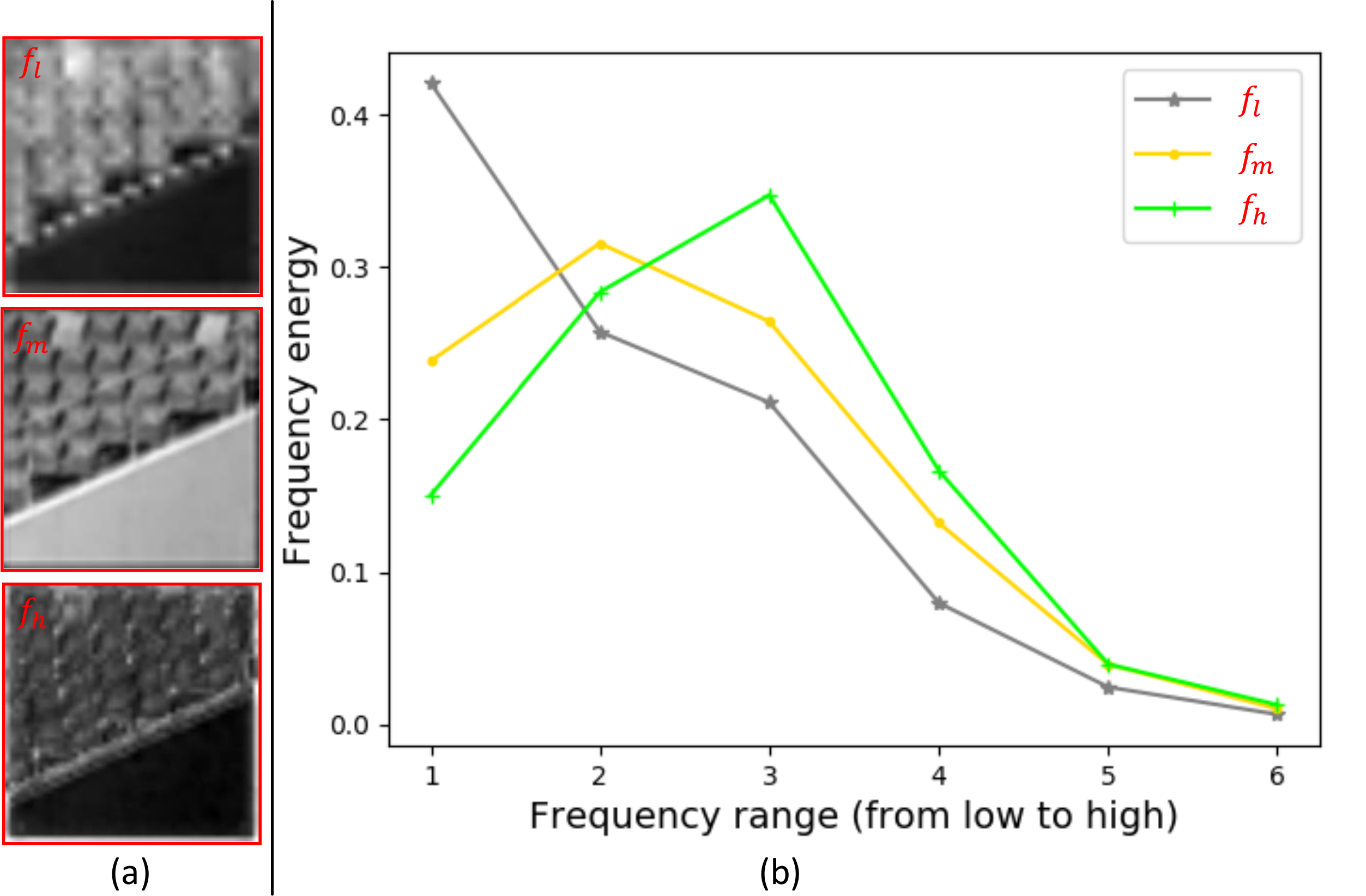}
	\caption{(a) Feature visualization and (b) frequency distribution analysis for FD module.}
	\label{fig:fd_vis}
	\vspace{-4mm}
\end{figure}

To study the influence of each module in our OR-Net, we visualize their inner features to understand how they work. For FD module, we first visualize the features for three frequency scales in Figure \ref{fig:fd_vis}(a), we see that the feature in high-frequency branch contains more details and texture information. We then analyse the low-/mid-/high-frequency features according to wavelet transform in Figure \ref{fig:fd_vis}(b). From left to right, with the frequency scale increases from 1 to 6, we find that the energy of low-frequency features $f_l$ is almost concentrated in the low frequency domain, and the energy of mid-/high-frequency features $f_m$/$f_h$ are concentrated in the relatively higher frequency domains, which validate the effectiveness of our frequency decomposition module.

\begin{figure}[htp]
	\centering
	\includegraphics[width=0.9\linewidth]{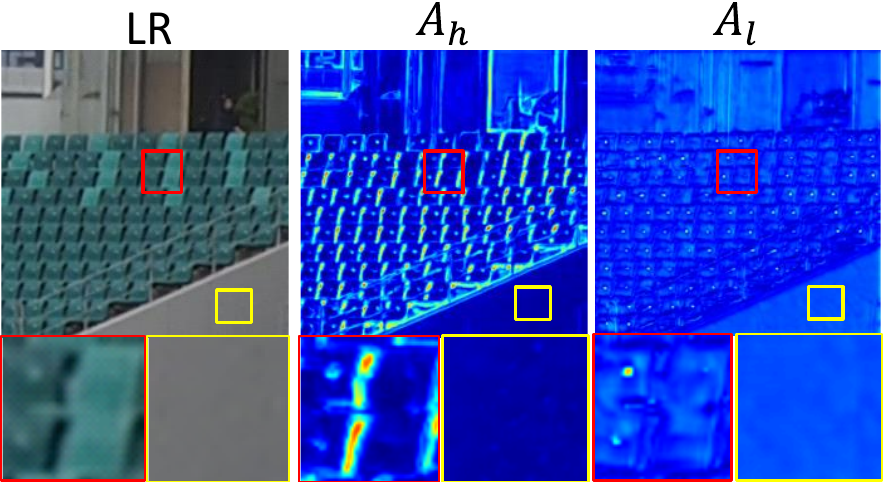}
	\vspace{-1mm}
	\caption{Visualization of the synthesized attention maps in RFA module, and $A_h, A_l \in A$.}
	\label{fig:rfa}
	\vspace{-1mm}
\end{figure}

For RFA module, we visualize the generated attention maps that used to adaptively aggregate frequency features in Figure \ref{fig:rfa}. We see that the attention map $A_h$ for high-frequency features (Note that $A_h \in A$) tends to enhance the high-frequency details at the object edges, and the low-frequency attention map $A_l$ ($A_l \in A$) tends to select the low-frequency features at the flat region. 

\begin{table}[htp]
\caption{Performance of different settings in OR-Net, where ``bran. = $i$'' means that there are $i$ decomposed frequency branches in FD module of OR-Net.}
\vspace{-2mm}
\setlength{\tabcolsep}{2.0mm}{
\begin{tabular}{c|c|c|c|c}
\toprule
Scheme & bran.=1 & bran.=2 & bran.=3 (ours) & bran.=4\\ \hline
PSNR    & 32.25         &  32.40        & \textbf{32.59}        & 32.52          \\ \hline
SSIM    & 0.856          &  0.859        & \textbf{0.877}      & 0.861         \\ \hline
LPIPS   & 0.310         &  0.299        & \textbf{0.230}      & 0.296         \\ \bottomrule
\end{tabular}}
\vspace{-2mm}
\label{tab:frequency}
\end{table}

\vspace{-2mm}

\subsection{Ablation Studies}\label{sec:ab}



\subsubsection{Effectiveness of omni-frequency.}
To study the influence of omni-frequency, we set the number of decomposed frequency branches in OR-Net as 1, 2, 3, 4 to get several schemes ( denoted as $bran.$ 1, 2,3, 4). As shown in Table \ref{tab:frequency}, we observe that the performance of OR-Net is improved as the number of frequency branch increases, but begins to decay when the number of branch exceeds 3. To balance the complexity and performance, we employ 3 frequency branches (corresponding to the low-/mid-/high-frequency components) in this paper.

\begin{table}[!htp]
	\centering
	\caption{Ablation experiments conducted on  DRealSR to study the effectiveness of the proposed RFA module and FEU design in our OR-Net.}
	\vspace{-2mm}
	\setlength{\tabcolsep}{4.5mm}{
		\begin{tabular}{cc|ccc}
			\toprule
			RFA & FEU & PSNR & SSIM & LPIPS\\ \hline
			$\checkmark$& $\checkmark$ & \textbf{32.59} & \textbf{0.863} & \textbf{0.292} \\ \hline
			$\times$& $\checkmark$ & 32.34  & 0.858 & 0.302\\
			$\checkmark$&$\times$ & 32.23  & 0.859  & 0.294  \\
			$\times$&$\times$   & 32.11   & 0.856  & 0.302 \\
			  SA & $\checkmark$   & 32.46 & 0.860 & 0.295 \\
			 \bottomrule
		\end{tabular}
	}
	\label{tab:ablation}
\end{table}

\vspace{-2mm}
\subsubsection{Effectiveness of our RFA module and FEU design.} As shown in Table. \ref{tab:ablation}, the proposed RFA module could bring 0.25dB gain in PSNR, and the FEU design also could bring 0.36dB gain in PSNR. Moreover, when we replace our RFA module with a general spatial attention (SA) to implement frequency aggregation, we see that our RFA module outperforms SA by 0.13dB in PSNR, which reveals that the dynamic convolution used in RFA promises better omni-frequency feature adaptive aggregation.
\vspace{-2mm}
\section{Conclusion}


In this paper, we propose an \textbf{O}mni-frequency \textbf{R}egion-adaptive \textbf{Net}work (OR-Net) to enable effective real image super-resolution. To efficiently promise the learned feature representations of OR-Net are both \emph{informative} and \emph{content-aware}, we first start from the frequency perspective and design a Frequency Decomposition (FD) module to fully leverage the omni-frequency features to comprehensively enhance texture details for LR images. Then, to adaptively aggregate such omni-frequency features to achieve the content-aware super-resolution, we further introduce a Region-adaptive Frequency Aggregation (RFA) module. Extensive experiments on several benchmarks demonstrate the superiority of OR-Net, and comprehensive ablation analysis  verify the effectiveness of FD and RFA modules.

\vspace{-2mm}
\section*{Acknowledgement} 
This work was supported in part by NSFC under Grant U1908209, 61632001 and the National Key Research and Development Program of China 2018AAA0101400.
\section{Appendix}
\subsection{Comprision with the State-of-the-Arts on RealSR dataset.}
In this section, we compare our OR-Net with the state-of-the-art traditional SISR models (including Bicubic, VDSR \cite{kim2016accurate}, EDSR \cite{lim2017enhanced}, RDN \cite{zhang2018residual}, DDBPN \cite{haris2018deep} and RCAN \cite{zhang2018image}), and the RealSR models (including LP-KPN \cite{cai2019toward} and CDC \cite{wei2020component})) on RealSR dataset, which is proposed by \cite{cai2019toward}. As shown in Table. \ref{tab:realsr}, we retrain all methods using RealSR dataset of $\times 2$. Compared with other methods, our OR-Net achieves best PSNR, SSIM, and LPIPS. For PSNR, we obtain a gain of 0.12dB compared with CDC \cite{wei2020component}.

\begin{table}[htp]
\caption{Quantitative results on the RealSR dataset of $\times 2$  \cite{wei2020component} comparison to the state-of-the-art methods in terms of PSNR, SSIM and LPIPS.}
\begin{tabular}{c|c|ccc}
\toprule
\multirow{2}{*}{Method} & \multirow{2}{*}{Category} & \multicolumn{3}{c}{RealSR} \\ \cline{3-5} 
                        &                           & PSNR    & SSIM    & LPIPS   \\ \midrule
Bicubic                 & \multirow{7}{*}{SISR}     & 31.67   & 0.887   & 0.223   \\
VDSR                    &                           & 32.14   & 0.871        &  0.197       \\
EDSR                    &                           & 33.88   & 0.920   & 0.145   \\
RDN                     &                           & 33.61   & 0.904   & 0.136        \\
DDBPN                   &                           & 33.53   & 0.898        & 0.145         \\
RCAN                    &                           & 33.83   & 0.923   & 0.147   \\ \midrule
LP-KPN                  &  \multirow{3}{*}{RealSR}                         & 33.90   & 0.927   & -       \\ 
CDC                     &    & 33.96   & 0.925   & 0.142   \\
OR-Net(Ours)            &                           & \textbf{34.08}   & \textbf{0.928}   & \textbf{0.127}   \\ \bottomrule
\end{tabular}
\label{tab:realsr}
\end{table}

\begin{table}[htp]
\caption{Quantitative results on the traditional super-resolution dataset, including Set5, Set14, and BSD100 of  $\times 2$  \cite{wei2020component} comparison to the state-of-the-art methods in terms of PSNR, and SSIM.}
\setlength{\tabcolsep}{0.8mm}{
\begin{tabular}{c|cc|cc|cc}
\toprule
\multirow{2}{*}{Method} & \multicolumn{2}{c|}{Set5} & \multicolumn{2}{c|}{Set14} & \multicolumn{2}{c}{BSD100}  \\ \cline{2-7} 
                        & PSNR        & SSIM        & PSNR        & SSIM         & PSNR         & SSIM                \\ \midrule
Bicubic                 & 33.66       & 0.9299      & 30.24       & 0.8688       & 29.56        & 0.8431            \\
SRCNN                   & 36.66       & 0.9542      & 32.45       & 0.9067       & 31.36        & 0.8879               \\
FSRCNN                  & 37.05       & 0.9560      & 32.66       & 0.9090       & 31.53        & 0.8920            \\
VDSR                    & 37.53       & 0.9590      & 33.05       & 0.9130       & 31.90        & 0.8960          \\
LapSRN                  & 37.52       & 0.9591      & 33.08       & 0.9130       & 31.08        & 0.8950            \\
MemNet                  & 37.78       & 0.9597      & 33.28       & 0.9142       & 32.08        & 0.8978            \\
EDSR                    & 38.11       & 0.9602      & \textbf{33.92}       & \textbf{0.9195}       & 32.32        & 0.9013           \\
MGBP                    & -           & -           & 33.27       & 0.9150       & 
31.99        & 0.8970          \\
SRMDNF                  & 37.79       & 0.9601      & 33.32       & 0.9159       & 32.05        & 0.8985             \\
DDBPN                   & 38.09       & 0.9600      & 33.85       & 0.9190       & 32.27        & 0.9000            \\ \midrule
OR-Net(ours)            & \textbf{38.12}            & \textbf{0.9610}            &  33.68           & 0.9178             & \textbf{32.33}            &  \textbf{0.9015}                        \\ \bottomrule
\end{tabular}}
\label{tab:generalsr}
\end{table}

\subsection{Comparison with the state-of-the-art methods on traditional super-resolution datasets.} 
With the development of deep learning, a series of studies \cite{ pan2020image, dong2015image, lim2017enhanced, dong2016accelerating} have achieved great progress on traditional super-resolution (SR) task.
In this section, we also verify our OR-Net on the traditional SR dataset(including Set5, Set14 and BSD100) of $\times 2$, that are down-sampled by bicubic. 
As shown in Table. \ref{tab:generalsr}, we compare our OR-Net with some state-of-the-art mothods, that are designed for traditional SR, including Bicubic, SRCNN \cite{dong2015image}, FSRCNN \cite{dong2016accelerating}, VDSR \cite{kim2016accurate}, LapSRN \cite{lai2018fast}, MemNet \cite{tai2017memnet}, EDSR \cite{lim2017enhanced}, MGBP \cite{michelini2019multigrid}, SRMDNF \cite{zhang2018learning}, and DDBPN \cite{haris2018deep}. 
We can also obtain performance comparable to EDSR and DDBPN. Above experiments verify that our OR-Net, which focus on solving omni-frequency degradation, is robustness to traditional SR, which exists the degradation in high frequency component. That is because the high frequency component is part of omni-frequency components.

\bibliography{ref.bib}

\end{document}